\documentclass{article} 
\usepackage{iclr2025_conference,times}

\usepackage{wrapfig}
\usepackage{lipsum} 
\usepackage{wrapfig}

\usepackage{booktabs}      
\usepackage{multirow}      
\usepackage{amssymb}       
\usepackage{graphicx, subfigure}     

\usepackage{colortbl}
\usepackage{xcolor}

\usepackage{hyperref}

\usepackage{url}
\usepackage{multicol}
\usepackage{bm}
\usepackage{aliascnt}
\usepackage{adjustbox}
\usepackage{siunitx} 
\usepackage{booktabs}
\usepackage{multirow} 
\usepackage{enumitem}
\usepackage{booktabs} 
\usepackage{amssymb}  
\usepackage{amsmath}  
\usepackage{graphicx} 
\usepackage{longtable} 
\usepackage{graphicx}
\usepackage{subcaption}
\usepackage{calc}
\usepackage[most]{tcolorbox}
\usepackage{listings}
\usepackage{wrapfig}
\usepackage{caption}

\definecolor{uclablue}{rgb}{0.15, 0.45, 0.68}
\hypersetup{
    breaklinks,
    citecolor=uclablue,
    colorlinks=true,
}

\newtcolorbox{AIbox}[2][]{aibox,title=#2,#1}
\tcbset{
  aibox/.style={
    top=10pt,
    colframe=black,
    colbacktitle=black,
    coltitle=white,
    center,
  }
}

\lstdefinelanguage{prompt}{
    basicstyle=\scriptsize\ttfamily, 
    mathescape=true,        
    escapebegin=\color{latentcolor},  
    escapeend={},
    escapechar=@,
    stringstyle = \color{myorange},
    showstringspaces = false,
    moredelim = [s][\color{mypink}]{`}{`},
    moredelim = [s][\color{mybrown}]{```json}{```},
    moredelim = [s][\color{latentcolor}]{<StartOfLatent>}{<EndOfLatent>},
    literate = %
        {\ \ a.\ }{{\textcolor{mypurple}{\ \ a.\ }}}5
        {\ \ b.\ }{{\textcolor{mypurple}{\ \ b.\ }}}5
        {\ \ c.\ }{{\textcolor{mypurple}{\ \ c.\ }}}5
        {\ \ d.\ }{{\textcolor{mypurple}{\ \ d.\ }}}5
        {\ \ e.\ }{{\textcolor{mypurple}{\ \ e.\ }}}5
        {\ \ f.\ }{{\textcolor{mypurple}{\ \ f.\ }}}5
        {\ \ g.\ }{{\textcolor{mypurple}{\ \ g.\ }}}5
        {\ \ h.\ }{{\textcolor{mypurple}{\ \ h.\ }}}5
        {\ I.\ }{{\textcolor{mypurple}{\ I.\ }}}4
        {\ II.\ }{{\textcolor{mypurple}{\ II.\ }}}5
        {\ III.\ }{{\textcolor{mypurple}{\ III.\ }}}6
        {\ IV.\ }{{\textcolor{mypurple}{\ IV.\ }}}5
        {\ V.\ }{{\textcolor{mypurple}{\ V.\ }}}4
}

\newtcblisting[auto counter, number within=subsection]{prompt}[4][]{%
  before skip=6pt, after skip=6pt,    
  top=6pt, bottom=6pt, left=7pt, right=7pt, 
  enhanced, breakable,
  colback=white, colframe=gray!65,
  boxrule=0.6pt, arc=2pt,
  drop shadow={black!15},             
  listing only,
  listing options={
    language=prompt,
    upquote=true,
    basicstyle=\scriptsize\ttfamily \setlength{\baselineskip}{1.1\baselineskip},
    columns=fullflexible,
    breaklines=true,
    breakindent=0pt,
    xleftmargin=\ifx\empty#2\empty0pt\else#2\fi,
    xrightmargin=\ifx\empty#3\empty0pt\else#3\fi,
    aboveskip=0pt,                    
    belowskip=0pt,                    
    showstringspaces=false,
  },
  title={\ifstrempty{#4}{Prompt \thetcbcounter}{Prompt \thetcbcounter: #4}},
  colbacktitle=gray!6, coltitle=black,
  attach boxed title to top left={yshift=-1mm, xshift=6pt},
  boxed title style={colback=gray!6, boxrule=0pt, sharp corners},
  #1
}

\newtcblisting[auto counter, number within=subsection]{showcase}[4][]{%
  before=\par\vspace{\baselineskip},
  after=\par,
  width=\linewidth,
  enhanced,
  arc=0em,
  boxrule=1pt,
  listing only,
  listing options={
    language=prompt,
    upquote=true,
    basicstyle=\scriptsize\ttfamily \setlength{\baselineskip}{1.1\baselineskip},
    breaklines=true,
    breakindent=0pt,
    xleftmargin=\ifx\empty#2\empty-12pt\else#2\fi,
    xrightmargin=\ifx\empty#3\empty-5pt\else#3\fi,
    aboveskip=-4pt,
    belowskip=-4pt,
    columns=fullflexible,
    },
  colback=white,
  colframe=gray,
  colbacktitle=gray!5,
  coltitle=black,
  attach boxed title to top center={yshift=-3mm},
  box align=center,
  parbox=false,
  title={\ifstrempty{#4}{Example \thetcbcounter}{Example \thetcbcounter: #4}},
  #1
}

\definecolor{linkColor}{rgb}{0.2,0.4,0.6}
\definecolor{myblue}{HTML}{0379AC}
\definecolor{myred}{HTML}{A50E50}
\definecolor{myorange}{RGB}{238, 133, 74}
\definecolor{latentcolor}{named}{cyan}
\definecolor{normalcolor}{RGB}{0, 0, 0}
\usepackage{marvosym}
\definecolor{lightblue1}{rgb}{0.97, 0.985, 1} 
\definecolor{lightblue2}{rgb}{0.92, 0.965, 1} 
\definecolor{lightblue3}{rgb}{0.84, 0.93, 1}
\definecolor{lightblue4}{rgb}{0.74, 0.87, 1}
\definecolor{lightblue5}{rgb}{0.64, 0.81, 1}
\definecolor{lightblue6}{rgb}{0.54, 0.75, 1}

\definecolor{lightgreen1}{rgb}{0.97, 1.00, 0.97}
\definecolor{lightgreen2}{rgb}{0.92, 0.98, 0.92}
\definecolor{lightgreen3}{rgb}{0.84, 0.95, 0.84}
\definecolor{lightgreen4}{rgb}{0.74, 0.91, 0.74}
\definecolor{lightgreen5}{rgb}{0.64, 0.86, 0.64}
\definecolor{lightgreen6}{rgb}{0.54, 0.81, 0.54}

\definecolor{lightorange1}{rgb}{1.00, 0.98, 0.95}
\definecolor{lightorange2}{rgb}{1.00, 0.95, 0.85}
\definecolor{lightorange3}{rgb}{1.00, 0.90, 0.70}
\definecolor{lightorange4}{rgb}{1.00, 0.85, 0.55}
\definecolor{lightorange5}{rgb}{1.00, 0.80, 0.40}
\definecolor{lightorange6}{rgb}{1.00, 0.75, 0.30}

\definecolor{lightpurple1}{rgb}{0.985, 0.97, 1.00}
\definecolor{lightpurple2}{rgb}{0.96, 0.92, 1.00}
\definecolor{lightpurple3}{rgb}{0.93, 0.84, 1.00}
\definecolor{lightpurple4}{rgb}{0.87, 0.74, 1.00}
\definecolor{lightpurple5}{rgb}{0.81, 0.64, 1.00}
\definecolor{lightpurple6}{rgb}{0.75, 0.54, 1.00}

\definecolor{lightred1}{rgb}{1.00, 0.97, 0.97}
\definecolor{lightred2}{rgb}{1.00, 0.92, 0.92}
\definecolor{lightred3}{rgb}{1.00, 0.84, 0.84}
\definecolor{lightred4}{rgb}{1.00, 0.74, 0.74}
\definecolor{lightred5}{rgb}{1.00, 0.64, 0.64}
\definecolor{lightred6}{rgb}{1.00, 0.54, 0.54}

\definecolor{lightcyan1}{rgb}{0.97, 1.00, 1.00}
\definecolor{lightcyan2}{rgb}{0.92, 0.98, 0.98}
\definecolor{lightcyan3}{rgb}{0.84, 0.95, 0.96}
\definecolor{lightcyan4}{rgb}{0.74, 0.91, 0.94}
\definecolor{lightcyan5}{rgb}{0.64, 0.87, 0.92}
\definecolor{lightcyan6}{rgb}{0.54, 0.83, 0.90}

\usepackage{xcolor,colortbl}
\definecolor{Gray}{gray}{0.85}
\definecolor{LightCyan}{rgb}{0.88,1,1}
\definecolor{greyC}{RGB}{180,180,180}
\definecolor{greyL}{RGB}{235,235,235}
\definecolor{citeColor}{RGB}{0,20,115}
\hypersetup{colorlinks,linkcolor={red},citecolor={citeColor},urlcolor={citeColor}}
\definecolor{shadecolor}{rgb}{0.92,0.92,0.92}

\usepackage{graphicx}
\usepackage{subcaption}
\usepackage{url}
\newcommand{\method}{iFSQ}

\usepackage{booktabs}
\usepackage{multirow}
\usepackage{cleveref}
\crefname{template}{Template}{Template}

\usepackage{enumitem} 
\usepackage[most]{tcolorbox}
\usepackage{pifont}
\definecolor{rliableblue}{RGB}{0, 102, 204} 

\newcommand{\sname}{iFSQ\xspace}
\tcbset{
    myblueboxstyle/.style={
        colback=lightblue3, 
        colframe=lightblue5, 
        coltitle=black, 
        fonttitle=\bfseries, 
        boxrule=1pt, 
        width=\linewidth, 
        left=2pt, 
        right=2pt, 
        top=2pt, 
        bottom=2pt, 
        sharp corners, 
        boxsep=2pt
    }
}

\lstdefinestyle{iclrstyle}{
    language=Python,
    basicstyle=\ttfamily\small,  
    columns=fullflexible,        
    keepspaces=true,             
    showspaces=false,            
    showstringspaces=false,      
    commentstyle=\color{gray}\itshape, 
    keywordstyle=\color{codekw}\bfseries, 
    stringstyle=\color{myorange}, 
    escapechar=|,                
    frame=none,                  
    xleftmargin=1.5em,           
    aboveskip=0.5em,             
    belowskip=0.5em,             
    breaklines=true,             
    breakindent=0pt,
}

\usepackage{algorithm}
\usepackage{algpseudocode}
\usepackage{listings} 
\usepackage{xcolor}   
\usepackage{etoolbox}
\makeatletter
\AfterEndEnvironment{algorithm}{\let\@algcomment\relax}
\AtEndEnvironment{algorithm}{\kern2pt\hrule\relax\vskip3pt\@algcomment}
\let\@algcomment\relax
\newcommand\algcomment[1]{\def\@algcomment{\footnotesize#1}}
\renewcommand\fs@ruled{\def\@fs@cfont{\bfseries}\let\@fs@capt\floatc@ruled
  \def\@fs@pre{\hrule height.8pt depth0pt \kern2pt}%
  \def\@fs@post{}%
  \def\@fs@mid{\kern2pt\hrule\kern2pt}%
  \let\@fs@iftopcapt\iftrue}
\makeatother

\NewDocumentCommand{\xx}
{ mO{} }{\textcolor{blue}{\textsuperscript{\textit{todo}}\textsf{\textbf{\small[#1]}}}}
\usepackage{xspace}

\usepackage{soul}
\definecolor{codeblue}{rgb}{0.25,0.5,0.5}
\definecolor{codekw}{rgb}{0.85, 0.18, 0.50}
\definecolor{diffgreen}{rgb}{0.0, 0.6, 0.0} 
\definecolor{diffred}{rgb}{0.8, 0.0, 0.0}   

\title{{\method}: Improving FSQ for Image Generation with 1 Line of Code}

\author{
Bin Lin$^{1,2}$, 
Zongjian Li$^{1}$, 
Yuwei Niu$^{1}$, 
Kaixiong Gong$^{2}$, 
Yunyang Ge$^{1,2}$, 
Yunlong Lin$^{2}$, 
Mingzhe Zheng$^{2}$,
JianWei Zhang$^{2}$, 
Miles Yang$^{2}$, 
Zhao Zhong$^{2}$, 
Liefeng Bo$^{2}$,
Li Yuan$^{1,\dagger}$\\
\textbf{$^1$Peking University}  \textbf{$^2$Tencent Hunyuan}\\
}

\begin{document}
\maketitle
\renewcommand*{\thefootnote}{\fnsymbol{footnote}}
\footnotetext{$*$ Work done during internship at Tencent Hunyuan. $\dagger$ Corresponding Authors.}

\begin{abstract} 
The field of image generation is currently bifurcated into autoregressive (AR) models operating on discrete tokens and diffusion models utilizing continuous latents. This divide, rooted in the distinction between VQ-VAEs and VAEs, hinders unified modeling and fair benchmarking. Finite Scalar Quantization (FSQ) offers a theoretical bridge, yet vanilla FSQ suffers from a critical flaw: its equal-interval quantization can cause activation collapse. This mismatch forces a trade-off between reconstruction fidelity and information efficiency. In this work, we resolve this dilemma by simply replacing the activation function in original FSQ with a distribution-matching mapping to enforce a uniform prior. Termed \textbf{\sname}, this simple strategy requires just \textbf{one line of code} yet mathematically guarantees both optimal bin utilization and reconstruction precision. Leveraging \sname as a controlled benchmark, we uncover two key insights: (1) The optimal equilibrium between discrete and continuous representations lies at approximately 4 bits per dimension. (2) Under identical reconstruction constraints, AR models exhibit rapid initial convergence, whereas diffusion models achieve a superior performance ceiling, suggesting that strict sequential ordering may limit the upper bounds of generation quality. Finally, we extend our analysis by adapting Representation Alignment (REPA) to AR models, yielding \textbf{LlamaGen-REPA}. Codes is available at \href{https://github.com/Tencent-Hunyuan/iFSQ}{https://github.com/Tencent-Hunyuan/iFSQ}.
\end{abstract}

\section{Introduction}

Image generation is currently bifurcated into two distinct paradigms: Autoregressive (AR) models that predict discrete image tokens (e.g., LlamaGen \cite{sun2024autoregressive}) and diffusion models that denoise continuous latent representations (e.g., DiT \cite{peebles2023scalable}). This divide is deeply rooted in their tokenizers: AR relies on Vector Quantized-VAEs (VQ-VAE) \cite{van2017neural} for discrete codes, while diffusion relies on Variational Autoencoders (VAE) \cite{kingma2013auto} for continuous distributions. This fragmentation creates a significant barrier to unified modeling and fair benchmarking. Specifically, it is difficult to disentangle whether performance differences stem from the generative models themselves (AR vs. Diffusion) or the distinct properties of their underlying tokenizers (VQ-VAE vs. VAE). To bridge this gap, we need a unified tokenizer capable of producing both high-quality discrete tokens and continuous latents.

Finite Scalar Quantization (FSQ) \cite{mentzer2023finite} emerges as a promising candidate. By replacing the complex learnable codebook of VQ-VAE with simple rounding operations, FSQ theoretically bridges the two worlds: its quantized values serve as continuous latents, while its rounded indices serve as discrete tokens. However, we identify a critical flaw when applying vanilla FSQ to visual generation: \textit{a mismatch between its equal-interval quantization and the non-uniform distribution of neural activations.}

\begin{figure*}[t]
\centering
    \includegraphics[width=1.0\linewidth]{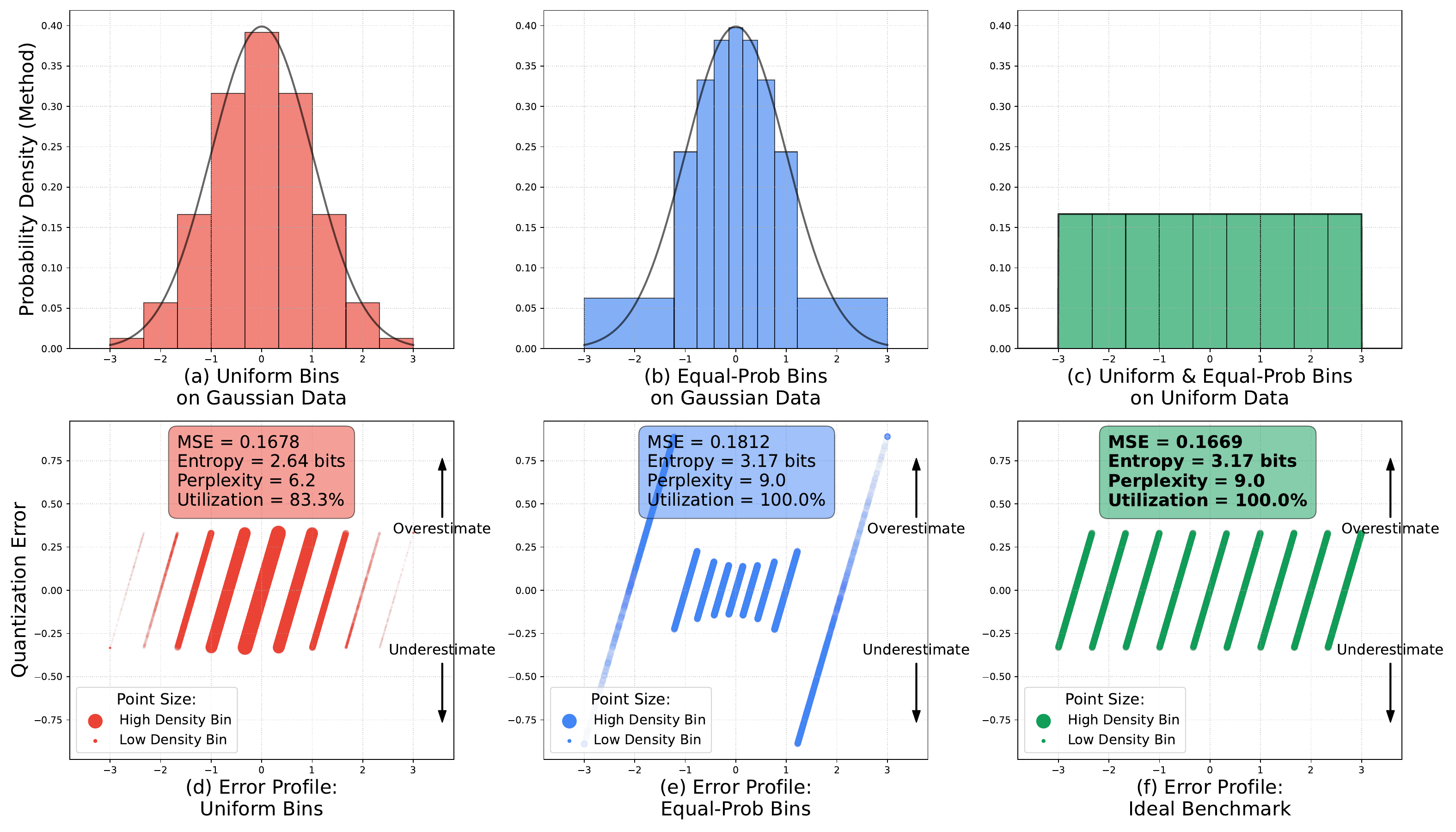}   
\caption{\textbf{Empirical analysis of equal-probability and equal-interval quantization.} Since neural network activations naturally follow a gaussian-like distribution \cite{lee2017deep}, we begin our experiments under this setting. For panels (a)–(f), we quantize the original data into 9 levels, which corresponds to an information entropy of 3.17 bits. We clip the original data to the range $[-3, 3]$ for visualization. Panel (a) shows equal-interval quantization of a standard normal distribution. Panel (b) shows equal-probability quantization of a standard normal distribution. Panel (c) shows equal-interval quantization of a uniform distribution, which is also equal-probability quantization. For panels (a)–(c), the area of each bin represents probability density. In panels (b) and (c), all bins have equal probability, while in panel (a), bins near the mean (0) have higher probability density. Panels (d)–(f) plot the quantization error for each original value (x-axis), where denser regions use larger markers.}
\label{fig:intro0}
\vskip -0.1in
\end{figure*}

As illustrated in \cref{fig:intro0}, this mismatch forces a trade-off between reconstruction fidelity and information efficiency: 
\begin{itemize} 
\vspace{-0.5mm}
\item \textbf{High Fidelity, Low Efficiency (\cref{fig:intro0} (a)):} Vanilla FSQ uses equal-interval bins. Since neural activations typically follow a bell-shaped (Gaussian-like) distribution, most data points crowd into the few central bins. While this dense sampling in the center yields low reconstruction error (MSE: 0.1678), lower reconstruction error is crucial for generating sharp images. However, it leaves edge bins severely underutilized. This ``activation collapse'' (Utilization: 83.3\%) limits the effective size, hindering to learn diverse patterns. 

\item \textbf{High Efficiency, Low Fidelity (\cref{fig:intro0} (b)):} Conversely, enforcing equal-probability for all bins maximizes information entropy (3.17 bits) and bin utilization (100\%). However, to accommodate the gaussian tails, the outer bins must be excessively wide. This coarse quantization at the edges leads to significant precision loss (MSE rises to 0.1812), degrading the visual quality of reconstructed images. 

\end{itemize}

In this work, we propose \textbf{\sname}, a tokenizer that unlocks the full potential of FSQ for generative modeling. Our key insight is simple yet effective: we replace the $tanh$ function in original FSQ with a distribution-matching activation that maps the unbounded gaussian latent space to a bounded uniform distribution. Specifically, the $tanh$ activation in original FSQ is replaced with $y = 2.0 \cdot \sigma(1.6 x) - 1$, which is \textbf{implemented in 1 line of code}. This activation ensures that the quantization bins are utilized with equal probability (high information efficiency) while maintaining equal intervals (high reconstruction fidelity). As shown in \cref{fig:intro0} (c), this elegant property guarantees optimal efficiency (100\% utilization) while simultaneously achieving superior fidelity (MSE: 0.1669).

Crucially, \sname establishes a fair and controlled benchmark for generative modeling. By using the exact same pre-trained tokenizer for both paradigms, we eliminate the confounding variables introduced by distinct discretization methods (e.g., VQ-VAE vs. VAE). Through this unified lens, we uncover two insights regarding the scaling properties of visual generation. First, we identify that the trade-off between discreteness and continuity has an optimal equilibrium appearing at approximately 4 bits per dimension. Second, when comparing paradigms under identical reconstruction constraints, we observe a distinct crossover: while AR models exhibit rapid convergence in early training, Diffusion models achieve a superior performance ceiling with sufficient compute. This suggests that the strict sequential inductive bias of AR may limit the upper bounds of generation quality compared to the holistic refinement of diffusion.

Beyond benchmarking, we adapt Representation Alignment (REPA) \cite{yu2024representation} to autoregressive models to further enhance their performance, yielding \textbf{LlamaGen-REPA}. We observe that aligning the 8-th layer of LlamaGen with visual features yields the best results, consistent with findings in diffusion models. Notably, AR models require significantly stronger alignment regularization ($\lambda=2.0$) compared to the standard $\lambda=0.5$ used in diffusion. 


Our contributions can be summarized in three aspects as follows:

\begin{itemize} 
\item \textbf{Methodology:} We propose a distribution-aware improvement to FSQ. By transforming gaussian latents into a uniform prior (implemented in 1 line of code), we resolve the conflict between information efficiency and reconstruction fidelity with a simple, plug-and-play activation function. 
\item \textbf{Benchmarking:} We introduce \sname as a unified tokenizer to benchmark AR against diffusion models. Our controlled experiments reveal that while AR excels in efficiency, diffusion dominates in peak generation quality, offering new guidance for model selection.
\item \textbf{Analysis \& Extension:} We conduct a comprehensive study on the quantization spectrum, identifying that a 4-bit representation serves as the ``sweet spot'' that balances the precision of continuous features with the compactness of discrete tokens. Furthermore, we successfully develop LlamaGen-REPA, identifying the critical role of alignment depth and loss weighting in balancing semantic alignment. 
\end{itemize}

\section{Related Work}



\subsection{Visual Tokenization: Continuous vs. Discrete}

Image generation faces a fundamental paradigm challenge, primarily bifurcated into diffusion models and autoregressive models. The core divergence lies in their choice of tokenizer: discrete versus continuous. Furthermore, in broader efforts to unify multi-modal architectures \cite{han2025vision,niu2025does,lin2025uniworld,li2025uniworld}, the design of the tokenizer remains unconverged.

\textbf{(1) Continuous:} Variational Autoencoders (VAEs) \cite{kingma2013auto} impose a gaussian prior on the latent space to support the probabilistic requirements of diffusion models. To ensure high reconstruction fidelity, recent works \cite{chen2025masked,yao2025reconstruction,li2025wf} incorporate adversarial losses (GAN) \cite{goodfellow2020generative}, perceptual losses (LPIPS) \cite{zhang2018unreasonable}, and even discriminative feature supervision (e.g., DINO series \cite{kang2023scaling,caron2021emerging,oquab2023dinov2,simeoni2025dinov3}) to enhance semantic alignment.

\textbf{(2) Discrete and explicit codebook}: AR models require discrete tokens, typically obtained via VQ-VAE \cite{van2017neural}. VQ-VAE quantizes latents by looking up the nearest neighbor in a learnable codebook. Despite its success, VQ-VAE suffers from codebook collapse and relies on the straight-through estimator for gradient approximation. While MAGVIT-v2 \cite{yu2023language} mitigates collapse via entropy regularization, the codebook lookup remains memory-intensive. Jukebox \cite{dhariwal2020jukebox} re-initializes underutilized codes by monitoring code usage statistics. In addition, pretrained codebook initialization is widely adopted, such as applying k-means clustering to encoder features \cite{zhu2024scaling} or initializing codebook from LLM embeddings \cite{han2025vision}.

\textbf{(3) Discrete but implicit codebook:} To circumvent the complexity and instability of learnable codebooks, Finite Scalar Quantization (FSQ) \cite{mentzer2023finite} explores scalar quantization strategies, which project latent representations directly onto a fixed, bounded grid via element-wise rounding. By eliminating the codebook lookup, FSQ simplifies the training dynamics and avoids codebook collapse. Moreover, FSQ is widely used for audio reconstruction (e.g., CoDiCodec \cite{pasini2025codicodec}, TTAE \cite{parker2024scaling} and Cosyvoice 2 \cite{du2024cosyvoice}), image reconstruction (e.g., MANZANO \cite{li2025manzano} and AToken \cite{lu2025atoken}) and video reconstruction (e.g., CS-FSQ \cite{argaw2025mambavideo}, Videoworld \cite{ren2025videoworld}, ViDTok \cite{tang2024vidtok} and Cosmos-discrete \cite{agarwal2025cosmos,liao2025langbridge}).

\subsection{Neural Network Quantization}

Quantization is widely adopted to reduce the memory footprint and computational cost of large models. In the realm of Large Language Models (LLMs) \cite{ahmed2025qwen,liu2024deepseek,achiam2023gpt}, Weight-only quantization methods like GPTQ \cite{frantar2022gptq} and AWQ \cite{lin2024awq} are prevalent for accelerating inference. Other approaches, such as QAT \cite{liu2024llm}, extend this to activation values. A core theoretical challenge in this field is Rate-Distortion Optimization (RDO) \cite{sullivan2002rate}, which seeks to minimize reconstruction error (Distortion) while minimizing the bit-rate (Rate) required for representation.
Difference: However, these methods typically treat quantization as a post-training compression step or an inference optimization technique. In contrast, our work integrates quantization directly into the tokenizer training phase, utilizing it as a fundamental mechanism to modulate the discreteness of the latent representation rather than merely for hardware acceleration.




\begin{figure*}[t]
\centering
    \includegraphics[width=1.0\linewidth]{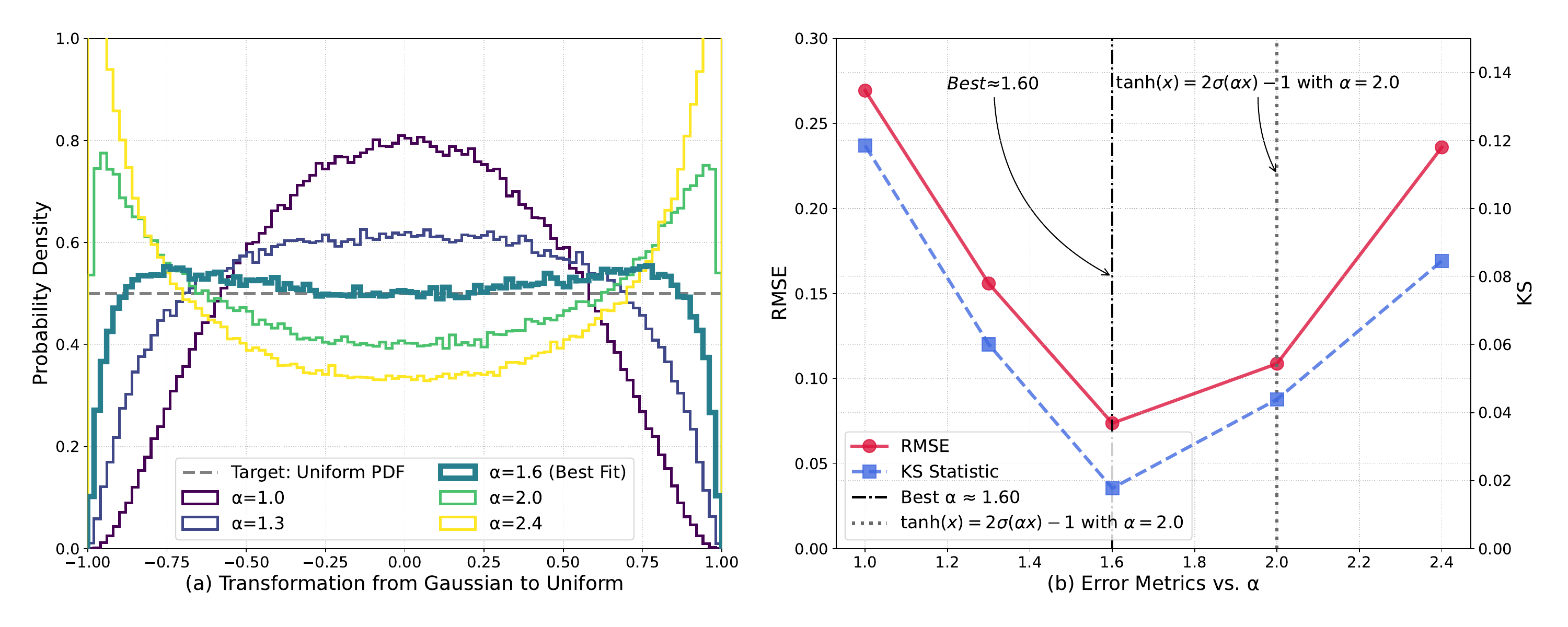}    
\caption{\textbf{Empirical numerical study of \bm{$2.0 \cdot \mathrm{sigmoid}(\alpha x) - 1.0$}.} Sample 500k points from the standard normal distribution and compute the transformed distribution for several values of $\alpha$. Panel (a) shows the probability density of $2.0 \cdot \mathrm{sigmoid}(\alpha x) - 1.0$ under different $\alpha$ values, and the panel (b) reports the similarity to a uniform distribution measured by KS and RMSE as $\alpha$ varies. Notably, the case with $tanh (\alpha = 2.0)$ corresponds to the original FSQ.}
\label{fig:method}
\end{figure*}

\section{\sname}
\label{sec:iFSQ}

\cref{sec:bg_fsq} outlines the FSQ pipeline, with additional context on autoregressive and diffusion tokenizers detailed in 
\cref{sec:bg_tok}. Subsequently, \cref{sec:ifsq} presents our approach to enhance FSQ via a one-line code change, validated by the qualitative and quantitative analyses in \cref{fig:method}.

\subsection{Background: Latent Quantization via FSQ}
\label{sec:bg_fsq}

We adopt a quantization workflow based on FSQ \cite{mentzer2023finite}. Given a latent representation $\bm{z} \in \mathbb{R}^{N \times d}$, we first apply a bounding function $f: \mathbb{R} \to [-1, 1]$ (typically $\tanh$) to constrain the value range. We define the quantization resolution with $L = 2^K + 1$ levels per channel, where the $+1$ term ensures the existence of an exact zero-center.

The continuous latent $\bm{z}$ is mapped to a vector of discrete integer indices $\bm{q} \in \{0, \dots, L-1\}^d$ through element-wise rounding. The quantization operation for the $j$-th dimension is formulated as:
\begin{equation}
    q_j = \text{round}\left( \frac{L-1}{2} \cdot (f(z_j) + 1) \right)
\end{equation}

This maps the range $[-1,1]$ to the integer set $\{0, \dots, L-1\}$.


\textbf{For Latent Diffusion Models (Continuous):} These models utilize the quantized feature vector $\bm{z}_{\text{quant}}$ directly as the input. We map the indices back to the continuous value space $[-1,1]$ via: 

\begin{equation} 
    z_{\text{quant}, j} = (q_j - \frac{L-1}{2}) \cdot \frac{2}{L-1} 
\end{equation} 

This operation acts as a lossy compression where $z_{\text{quant}} \approx z$.


\textbf{For Autoregressive Models (Discrete):} These models require a flat token representation. Unlike VQ-VAE which requires a learnable codebook lookup, FSQ projects the vector of indices $\bm{q}$ to a single scalar index $I$ via a bijective base-$I$ expansion: 
\begin{equation} 
    I = \sum_{j=1}^{d} q_j \cdot L^{d-j} 
\end{equation} 

\textbf{Example:} Consider a 4-dimensional vector $z$ with $K=1$ (implying $L=3$). If the quantization yields $\bm{q} = [2, 2, 1, 0]$, the unique codebook index is: \begin{equation*} I = 2 \cdot 3^3 + 2 \cdot 3^2 + 1 \cdot 3^1 + 0 \cdot 3^0 = 75 \end{equation*} 
The implicit codebook size is $|\mathcal{C}| = L^d = (2^K + 1)^d$.

\subsection{Distribution Analysis and Optimization for \sname}
\label{sec:ifsq}

\begin{algorithm}[t]
\caption{Pseudocode of \sname in a PyTorch-like style}
\algcomment{\fontsize{7.2pt}{0em}\selectfont \texttt{sigmoid}: logistic sigmoid function; \texttt{round}: element-wise rounding.
\vspace{-1mm} 
}

\begin{lstlisting}[style=iclrstyle]
def iFSQ(z, levels):
    '''
    z: visual feature map (B*H*W, D)
    levels: list or tensor defining levels per dim (L)
    '''
    
    # 1. Bound input to [-1, 1]
    # We replace the tanh to achieve a more uniform distribution.
    |\color{diffred}-\quad z = tanh(z)|
    |\color{diffgreen}+\quad z = 2 * sigmoid(1.6 * z) - 1|

    # 2. Scale to the grid defined by levels
    # half_width corresponds to (L-1)/2 in Eq. (3)
    half_width = (levels - 1) / 2
    z_scaled = z * half_width

    # 3. Quantization with Straight-Through Estimator
    z_rounded = round(z_scaled)
    z_hat = z_rounded - z_scaled.detach() + z_scaled

    # 4. Normalization for diffusion in Eq. (4)
    z_q = z_hat / half_width

    # 5. Compute Indices for AR in Eq. (5)
    # basis: [L^(d-1), ..., L^0]
    z_ind = z_rounded + half_width 
    basis = compute_basis(levels) 
    indices = sum(z_ind * basis, dim=-1).long() 

    return z_q, indices
\end{lstlisting}
\label{algo:ifsq}
\end{algorithm}

Clearly, for both diffusion and autoregressive models, the input feature distribution is intrinsically linked to the activation function $f(z)$. Typically, model features follow a normal distribution. However, passing a normal distribution through the $\tanh$ activation (employed in the original FSQ) yields a non-uniform, bimodal distribution, as illustrated by the green curve in \cref{fig:method} (a).

Furthermore, we investigate the general form of the sigmoid function:
\begin{equation}
    s(x) = A \cdot \sigma(\alpha x) + B
\end{equation}
where setting $A=2.0$, $\alpha=2.0$, and $B=-1$ renders it equivalent to $\tanh$. Consequently, we sweep the parameter $\alpha$ to determine if a specific value transforms the standard normal distribution $s(x)$ into an approximate uniform distribution.


\textbf{Qualitative Analysis:} As shown in \cref{fig:method} (a), we visualize the probability density functions (PDF) for $\alpha \in \{1.0, 1.3, 1.6, 2.0, 2.4\}$. As $\alpha$ transitions from 1.0 to 1.6, the distribution shifts from a symmetric unimodal shape to a uniform distribution. Conversely, as $\alpha$ increases from 1.6 to 2.4, the distribution becomes concave, forming a bimodal structure. We observe that $\alpha=1.6$ (dark green solid line) most closely approximates the uniform distribution (grey dashed line).

\textbf{Quantitative Analysis:} As depicted in \cref{fig:method} (b), we employ the Root Mean Square Error (RMSE) \cite{gauss1877theoria} and the Kolmogorov-Smirnov (KS) \cite{an1933sulla} statistic to quantify the proximity of the post-activation distribution to a uniform distribution.

\noindent \textbf{RMSE:} This metric measures the deviation between the quantized and target values.
\begin{equation}
    \text{RMSE} = \sqrt{ \frac{1}{N} \sum_{i=1}^{N} (x_i - \hat{x}_i)^2 }
\end{equation}
where $N$ is the number of samples, $x_i$ denotes the target value (ideal uniform), and $\hat{x}_i$ denotes the activated data point.

\noindent \textbf{KS:} This statistic measures the maximum divergence between two probability distributions (e.g., the empirical distribution $P$ and the target distribution $Q$). It is defined as the supremum of the absolute difference between their Cumulative Distribution Functions (CDFs):
\begin{equation}
    D_{KS} = \sup_{x} | F_{P}(x) - F_{Q}(x) |
\end{equation}

Corroborating our qualitative findings, the quantitative metrics reveal a clear trend: both RMSE and KS statistics reach their minima at $\alpha=1.6$, significantly outperforming the standard $\tanh$ baseline ($\alpha=2.0$). This confirms that $\alpha=1.6$ effectively transforms the gaussian input into a near-uniform distribution within the bounded range. Theoretically, a uniform distribution is optimal for the fixed-interval quantization employed by FSQ. It ensures that the static quantization bins are utilized with equal probability, thereby maximizing information entropy and mitigating the activation collapse observed in the original design. We designate this optimized formulation as \textbf{\sname} and empirically demonstrate in \cref{sec:t1_recon} that this distributional alignment translates directly to superior image reconstruction quality.



\cref{algo:ifsq} presents the pseudo-code for \sname. Our method is remarkably straightforward, requiring only a minimal modification to the activation function within the standard FSQ framework. Specifically, we replace the original bounding function, equivalent to $2\sigma(2z) - 1$ (i.e., $tanh(z)$), with the optimized form $2\sigma(1.6z) - 1$. This adjustment introduces no additional parameters or inference latency, serving as a computationally free, plug-and-play module compatible with existing architectures.

\begin{figure*}[t]
\centering
    \includegraphics[width=1.0\linewidth]{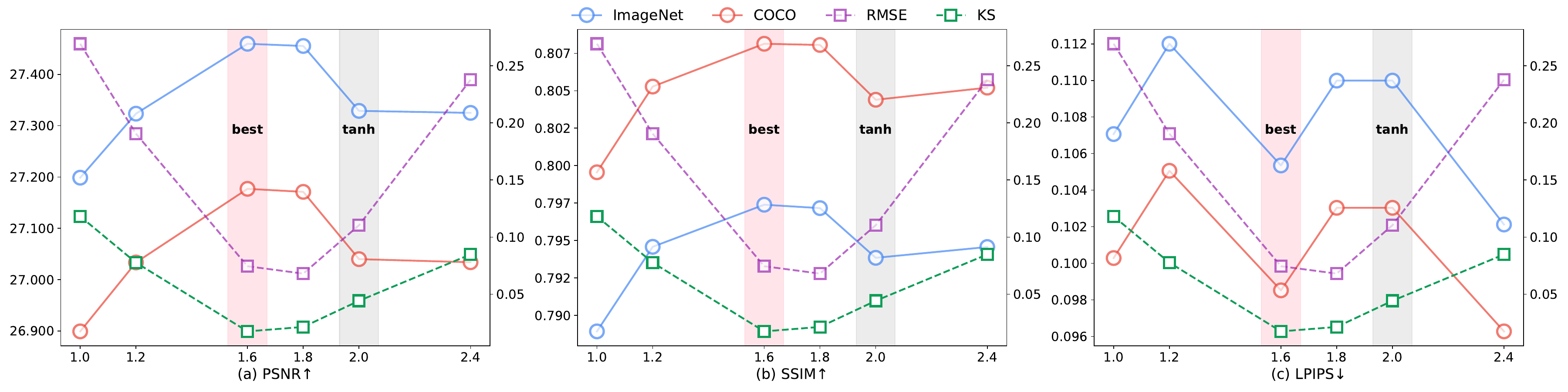}    
\caption{\sethlcolor{orange!15}\hl{\textbf{Effect of $\alpha$ for \sname.}} In (a)–(c), the x-axis denotes $\alpha$. The primary y-axes show PSNR (↑ better), SSIM (↑ better), and LPIPS (↓ better), respectively. The secondary y-axis shows distribution metrics RMSE and KS (both ↓ better). The optimal choice at $\alpha = 1.6$ is highlighted, and tanh performance at $\alpha = 2.0$ is marked, which corresponds to the original FSQ.}
\label{fig:tanhscale}
\end{figure*}

\section{Experiments}
\label{sec:exp}

We validate the effectiveness of \sname through extensive experiments in \cref{sec:exp_ifsq}. Additionally, we extend REPA to LlamaGen in \cref{sec:exp_llamagen_repa}.

\subsection{Experiments for \sname}
\label{sec:exp_ifsq}

Specifically, we investigate the following questions for \sname and show the setup of tokenzier in \cref{sec:setup}: 
\begin{itemize}[leftmargin=*,itemsep=0mm]
\item \textbf{Methodology:} For the image reconstruction and generation performance, does the latent representation of images encoded by \sname outperform discrete VQ-VAE, FSQ and continuous AE? (\cref{tab:dit_wo_repa},  \cref{tab:llamagen_res})
\item \textbf{Benchmarking:} Which performs better for image generation on a unified tokenizer benchmark: autoregressive models or diffusion models? (\cref{fig:dit_llamagen})
\item \textbf{Analysis:} Can \sname achieve a better balance between discrete and continuous reconstruction performance? (\cref{tab:tok_baseline}, \cref{fig:quant_level}, \cref{fig:compress_ratio}). 
\end{itemize}

\subsubsection{\sname for Image Reconstruction}
\label{sec:t1_recon}

\sethlcolor{orange!15}
\hl{\textbf{Comparison between FSQ and \sname.}} In \cref{fig:tanhscale}, plot performance curves for $\alpha = \{1.0, 1.2, 1.6, 1.8, 2.0, 2.4\}$ with KS and RMSE overlaid. Notably, the case with $\alpha=2.0$ corresponds to the original FSQ. As $\alpha$ increases from 1.0 to 1.6, PSNR and SSIM increase while KS and RMSE decrease. We observe that iFSQ ($\alpha=1.6$) consistently outperforms the original FSQ ($\alpha=2.0$) across PSNR, SSIM, and LPIPS. As $\alpha$ increases from 1.6 to 2.4, KS and RMSE increase while PSNR and SSIM decrease. This pattern matches the analysis in the \cref{sec:ifsq}: at $\alpha = 1.6$, KS and RMSE reach minima and theoretical Efficiency–Fidelity trade-offs are optimal. LPIPS attains its best at $\alpha = 2.4$, but considering PSNR and SSIM, choose $\alpha = 1.6$ as the optimal setting. Moreover, although training is conducted only on ImageNet, similar trends are observed on the COCO validation set, demonstrating the strong scalability and robustness of \sname.

\subsubsection{\sname for Diffusion Image Generation}
\label{sec:t1_dit}

\sethlcolor{yellow!30}\hl{\textbf{Comparison between AE, FSQ and \sname.}} As shown in \cref{tab:dit_wo_repa}, we do not use REPA for training DiT. Using \sname as the tokenizer yields a better gFID (12.76) than AE (13.78), while \sname achieves a 3$\times$ higher compression rate (96 vs. 24). The same trend appears in REPA, where \sname at 4 bits already reaches generation performance comparable to AE.

\sethlcolor{violet!15}\hl{\textbf{Comparison of different bits within \sname.}} In \cref{tab:dit_wo_repa}, \sname at 2 bits yields substantially worse gFID than AE (18.52 vs. 13.78 without REPA; 14.97 vs. 10.67 with REPA). As quantization level increases to 4 bits, \sname performance becomes comparable to AE. For bit $>$ 4, \sname shows no consistent improvement and instead fluctuates. \sname at 5–8 bits does not differ markedly from AE. This pattern suggests a 4-bit latent already captures most features, and enlarging the latent space does not necessarily yield faster convergence.

\subsubsection{\sname for Auto-regressive Image Generation}
\label{sec:t1_llamagen}

\sethlcolor{yellow!30}\hl{\textbf{Comparison between VQ, FSQ and \sname.}} To accelerate experiments, we use LlamaGen-REPA for both \sname and VQ in \cref{tab:llamagen_res} with 256$\times$ spatial compression (introduced the background of the compression ratio in \cref{sec:compression}.) following the original LlamaGen. At the same latent dimension, autoregressive generation trained on VQ underperforms \sname, while \sname operates at a lower bit rate. 

\sethlcolor{violet!15}\hl{\textbf{Comparison of different bits within \sname.}} Larger bits (and thus larger codebooks) do not necessarily yield better results and performance peaks at 4 bits. Conjecture that as the codebook grows, the corresponding autoregressive model must also scale to provide sufficient capacity to predict such a large codebook.

\begin{table}[t]
    \centering
    \begin{minipage}[t]{0.49\textwidth}
        \caption{\textbf{FID comparison on DiT-Large without CFG.} All metrics evaluate on the ImageNet validation set. For AE, ``16 bit'' denotes using 16-bit precision for inference. ↓ indicates lower is better. Iter. denotes training iterations. Numbers in ``()'' indicate with REPA performance. CR denotes the compression ratio.}
        \label{tab:dit_wo_repa}
        \centering
        \resizebox{\linewidth}{!}{
          \begin{tabular}{c|ccc|ccc}
            \toprule
            \textbf{Tokenizer} & \textbf{Model} & \textbf{CR} & \textbf{Bit} & \textbf{Iter.} & \textbf{gFID↓} \\
            \midrule
            \cellcolor{yellow!30}AE & DiT-L/2 & 24 & 16 & 100k & 13.78 (10.67) \\
            \cellcolor{yellow!30}FSQ & DiT-L/2 & 96 & 4 & 100k & 13.38 (11.04) \\
            \cellcolor{yellow!30}\sname & DiT-L/2 & \textbf{96} & \textbf{4} & 100k & \textbf{12.76 (10.48)} \\
            \midrule
            \multirow{7}{*}{\sname} & DiT-L/2 & 192 & \cellcolor{violet!15}2 & 100k & 18.52 (14.97) \\
             & DiT-L/2 & 128 & \cellcolor{violet!15}3 & 100k & 14.51 (11.74) \\
             & DiT-L/2 & \textbf{96} & \cellcolor{violet!15}\textbf{4} & 100k & \textbf{12.76 (10.48)} \\
             & DiT-L/2 & 76 & \cellcolor{violet!15}5 & 100k & 14.35 (10.77) \\
             & DiT-L/2 & 64 & \cellcolor{violet!15}6 & 100k & 15.02 (10.74) \\
             & DiT-L/2 & 54 & \cellcolor{violet!15}7 & 100k & 12.80 (10.51) \\
             & DiT-L/2 & 48 & \cellcolor{violet!15}8 & 100k & 14.06 (10.54) \\
            \bottomrule
          \end{tabular}
        }
    \end{minipage}
    \hfill 
    \begin{minipage}[t]{0.49\textwidth}
        \caption{\textbf{FID comparison on LlamaGen-Large without CFG..} All metrics evaluate on the ImageNet validation set and we report gFID with LlamaGen-REPA. ↓ indicates lower is better. Iter. denotes training iterations. The 14 bit VQ donotes the LlamaGen-Large trains with 16,384 codebook size.}
        \label{tab:llamagen_res}
        \centering
        \resizebox{\linewidth}{!}{
          \begin{tabular}{c|ccc|cc}
            \toprule
            \textbf{Tokenizer} & \textbf{Model} & \textbf{Dim} & \textbf{Bit} & \textbf{Iter.} & \textbf{gFID↓} \\
            \midrule
            \cellcolor{yellow!30}VQ & LlamaGen-L & 4 & 14 & 500k & 33.90 \\
            \cellcolor{yellow!30}FSQ & LlamaGen-L & 4 & 2 & 500k & 32.48 \\
            \cellcolor{yellow!30}\sname & LlamaGen-L & 4 & 2 & 500k & 31.09 \\
            \cellcolor{yellow!30}VQ & LlamaGen-L & 8 & 14 & 500k & 29.91 \\
            \cellcolor{yellow!30}\sname & LlamaGen-L & 8 & 2 & 500k & \textbf{26.02} \\
            \midrule
            \sname & LlamaGen-L & 4 & \cellcolor{violet!15}3 & 500k & 29.02 \\
            \sname & LlamaGen-L & 4 & \cellcolor{violet!15}\textbf{4} & 500k & \textbf{28.07} \\
            \sname & LlamaGen-L & 4 & \cellcolor{violet!15}5 & 500k & 29.12 \\
            \sname & LlamaGen-L & 4 & \cellcolor{violet!15}6 & 500k & 32.60 \\
            \bottomrule
          \end{tabular}     
        }
    \end{minipage}
\end{table}

\subsubsection{Training Efficiency Comparison} 

\begin{wrapfigure}{r}{0.55\linewidth}
\vskip -0.15in
\centering
    \includegraphics[width=1.0\linewidth]{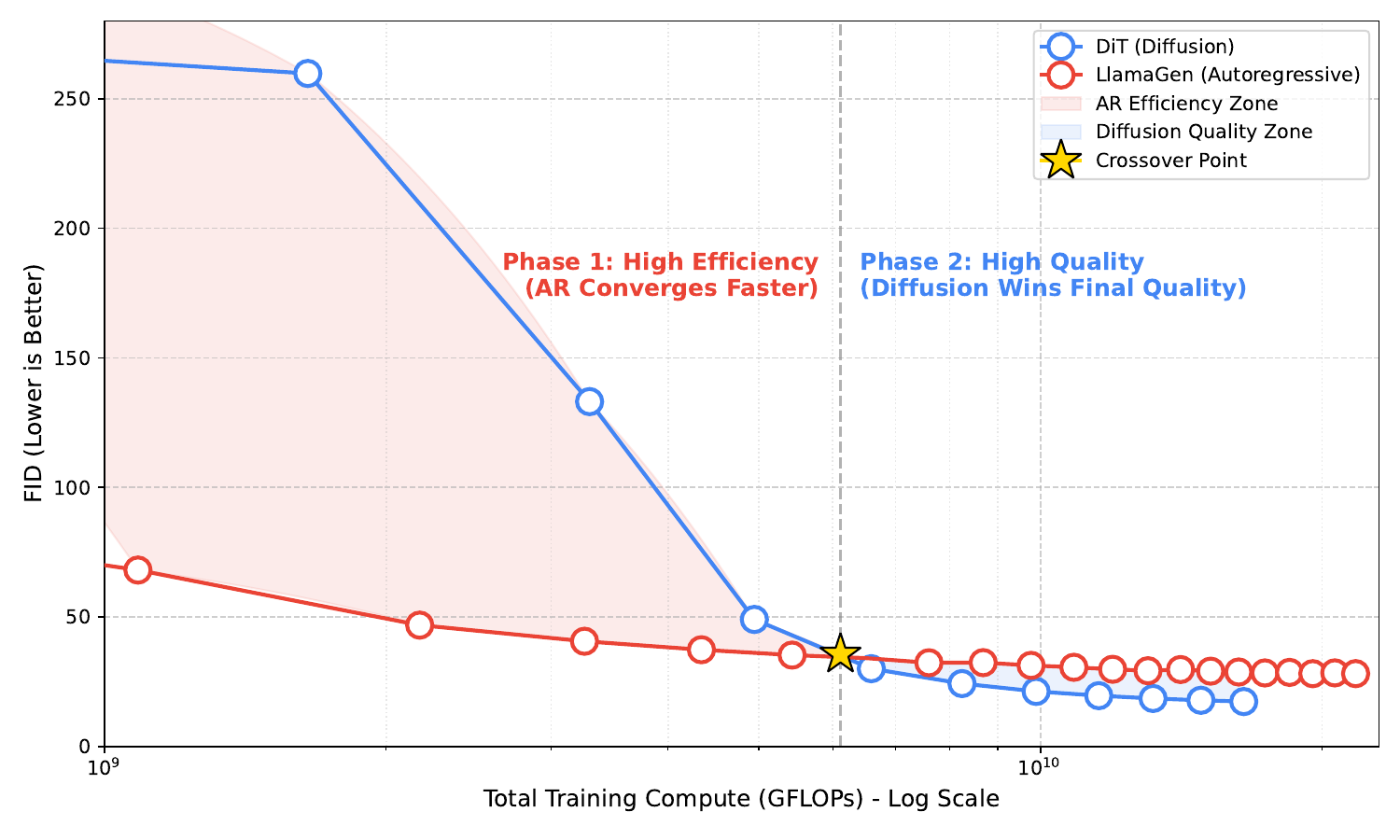}
\caption{\textbf{Training Efficiency Comparison: DiT vs LlamaGen (FID vs Compute).} At 256 resolution, DiT-Large and LlamaGen-L exhibit approximately 161.04G and 169.65G FLOPs, respectively. Both models employ optimal training configurations derived from ablation studies while sharing the same \sname.}
\label{fig:dit_llamagen}
\vskip -0.1in
\end{wrapfigure}
As demonstrated in \cref{sec:t1_dit} and \cref{sec:t1_llamagen}, \sname achieves performance comparable to AE and VQ-VAE in diffusion and autoregressive generation, respectively. Since \sname as both a continuous latent and a discrete index, it establishes a fair platform for comparing diffusion and AR models by the same decoder reconstruction performance. As illustrated in \cref{fig:dit_llamagen}, we plot the performance scaling against computational resources for both models using the same tokenizer to investigate their training efficiency. For LlamaGen, the FLOPs for attention computation and value weighting are calculated as half of full attention. We observe that while diffusion models exhibit slower initial convergence compared to the rapid convergence of AR models, they eventually reach a crossover point. Beyond this, diffusion models continue to achieve superior performance, whereas AR models show limited gains. This suggests that the strong sequential constraint is suboptimal for image generation.

\subsubsection{Scaling Behavior of \sname}
\label{sec:scale_ifsq}

\begin{figure*}[!t]
\centering
    \includegraphics[width=1.0\linewidth]{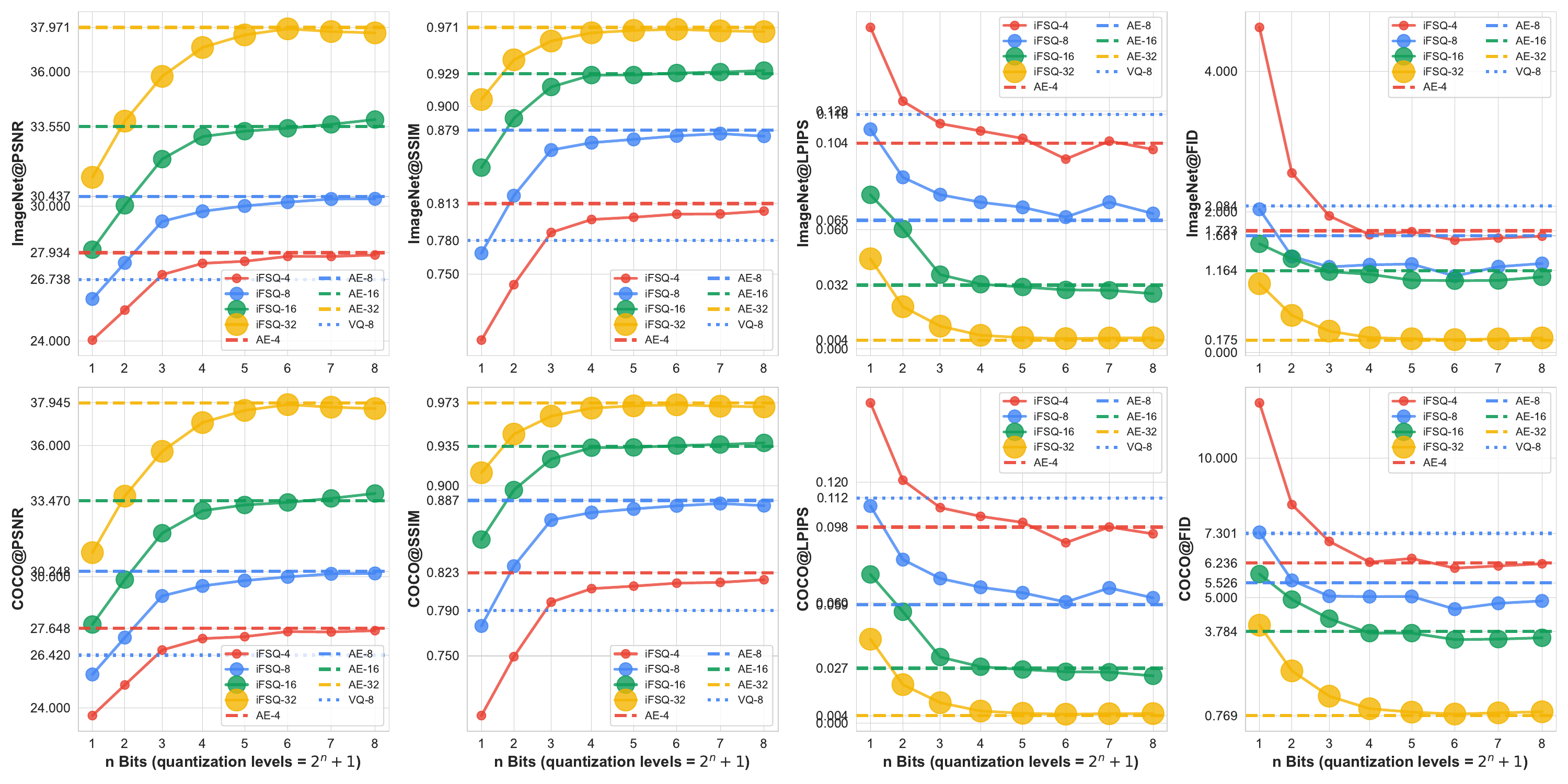}
\caption{\textbf{Performance across quantization levels.} We plot the performance of \sname and AE under different quantization levels, using larger markers to denote models with higher latent dimensionality. Each performance point (including AE) uses a spatial compression factor of 64$\times$. The performance of AE is indicated by horizontal dashed lines, which train under mixed precision and use 16-bit precision to inference.}
\label{fig:quant_level}
\end{figure*}

As shown in \cref{fig:quant_level}, \sname achieves a favorable trade-off across quantization levels: at high quantization levels performance approaches AE, while at low quantization levels performance remains substantially stronger than VQ. PSNR, SSIM, LPIPS, and FID are reported on ImageNet and COCO. All metrics exhibit similar trends: (1) at the same latent dimension, \sname performance improves as quantization level increases. \sname approaches AE around 4 bits, is nearly identical to AE at 7–8 bits, and \sname with 16-dim attains PSNR and SSIM that exceed AE. (2) at the same quantization level, larger latent dimensions improve performance for both \sname and AE. Notably, VQ-8dim performs worse than \sname-4dim or AE-4dim, suggesting that learning a quantization scalar is easier than learning quantization embeddings. We also observe that at 2 bits, \sname with twice the latent dimension (\sname-2$x$dim) already surpasses AE at $x$-dim. This trend holds across quantization levels and latent dimensions, indicating strong scalability of \sname. Next, we analyze scalability further using compression ratio in \cref{sec:res_cr}.

\subsection{Experiments for LlamaGen-REPA}
\label{sec:exp_llamagen_repa}

\begin{figure*}[!t]
\centering
    \includegraphics[width=1.0\linewidth]{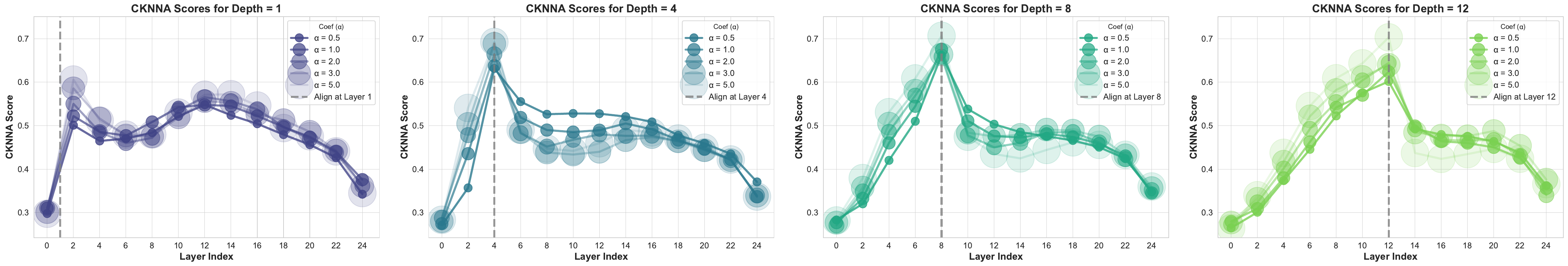} 
\caption{\textbf{Layer-wise CKNNA scores under different REPA alignment depths (1, 4, 8, and 12).} The vertical dashed lines denote the alignment layers, and bubble sizes correspond to the training coefficient $\lambda$. The peak semantic alignment consistently synchronizes with the target alignment depth.}
\label{fig:cknna}
\end{figure*}

Specifically, we investigate the internal dynamics of autoregressive models and the optimal configuration for representation alignment. We address the following questions:

\begin{itemize}[leftmargin=*,itemsep=0mm]
\item \textbf{Quantitative Analysis:} How do feature representations evolve layer-by-layer in autoregressive models? Do they exhibit a clear transition from self-encoding to next-token prediction? (\cref{fig:sim_cknna})
\item \textbf{Semantic Acceleration:} Can explicit feature alignment (REPA) effectively guide the model to acquire high-level semantics at earlier layers? (\cref{fig:cknna})
\item \textbf{Ablation \& Scaling:} What are the optimal hyperparameters (target representation, alignment depth, and loss coefficient) for LlamaGen-REPA? Does the optimal alignment depth generalize across different model scales? (\cref{tab:llamagen_repa_abla}, \cref{fig:llamagen_dit_depth_repa}, \cref{fig:coef_repa_llamagen})
\end{itemize}

\subsubsection{Quantitative Analysis}

To elucidate the internal dynamics of autoregressive image generation models, we introduce three metrics to quantify layer-wise feature evolution.

First, \textbf{Self-Token Similarity (STS)} measures the retention of input encoding by calculating the spatially aligned cosine similarity between the feature at layer $l$, denoted as $h_l$, and the input embedding $h_0$:
\begin{equation}
    \text{STS}_l = \frac{1}{N} \sum_{i=1}^{N} \cos(h_l^{(i)}, h_0^{(i)}),
\end{equation}
where $N$ represents the total number of tokens and $h^{(i)}$ denotes the feature at spatial position $i$.

Second, to identify the transition to a prediction state, we define \textbf{Next-Token Similarity (NTS)}. The core mechanism of autoregressive modeling dictates that as the network deepens, the feature at the current position $i$ must evolve to predict the content of the next position $i+1$. Therefore, unlike STS which compares aligned positions, NTS measures the \textit{shifted} similarity between the current layer's features (from index $1$ to $N-1$) and the input embedding (from index $2$ to $N$):
\begin{equation}
    \text{NTS}_l = \frac{1}{N-1} \sum_{i=1}^{N-1} \cos(h_l^{(i)}, h_0^{(i+1)}).
\end{equation}
A rise in NTS indicates that the layer has shifted focus from encoding the current token to predict the subsequent token, signaling the onset of the generation mode.

Third, \textbf{CKNNA} assesses the alignment between the global semantics of the current layer and pre-trained DINOv2 features. Following the protocol in REPA \cite{yu2024representation}, we utilize CKNNA score to quantify this semantic correspondence without defining a new formula.

\begin{figure*}[!t]
\centering
    \includegraphics[width=1.0\linewidth]{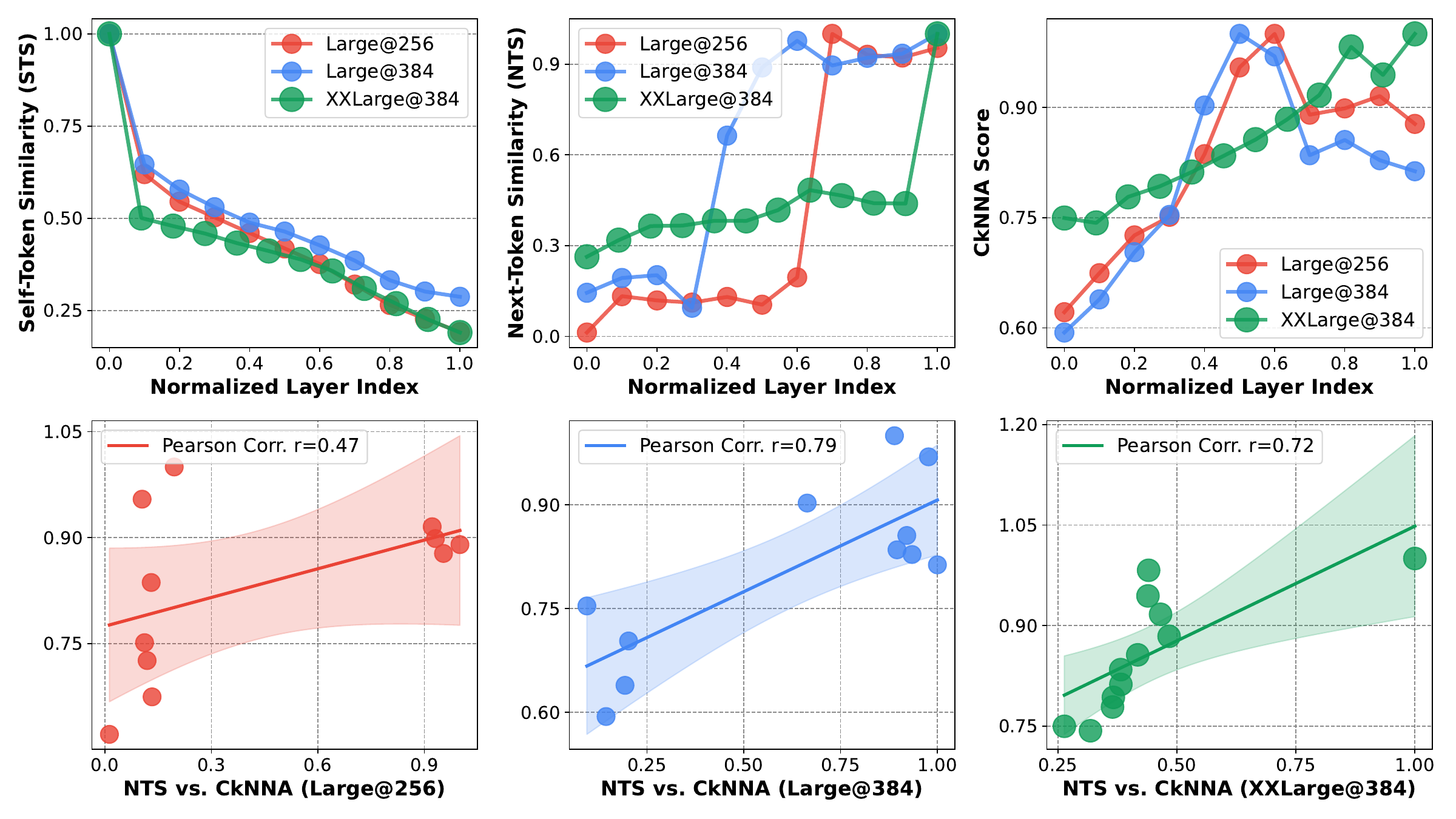} 
\caption{\textbf{Layer-wise analysis of LlamaGen models.} The top row shows the evolution of STS, NTS, and CKNNA across normalized layer indices. The bottom row demonstrates the strong correlation between NTS and CKNNA, particularly at higher resolutions and model scales. The shaded regions indicate the 90\% confidence intervals.}
\label{fig:sim_cknna}
\end{figure*}

\cref{fig:sim_cknna} illustrates the qualitative and quantitative analysis of these metrics across different model scales (Large, XXLarge) and resolutions (256, 384). The top row reveals distinct trends: STS decreases as network depth increases, indicating a gradual departure from the initial self-encoding state. Conversely, both NTS and CKNNA exhibit a sharp increase in the middle-to-late layers. Notably, the layer index where NTS surges—signaling the onset of the prediction mode—synchronizes highly with the rise in CKNNA.

To rigorously validate this relationship, the bottom row of \cref{fig:sim_cknna} presents the linear fit between NTS and CKNNA. We observe a strong positive correlation, with Pearson coefficients of $r=0.47$ (Large@256), $r=0.79$ (Large@384), and $r=0.72$ (XXLarge@384). This correlation becomes more pronounced as model scale and resolution increase, suggesting that the emergence of the \textit{next-token prediction} capability is intrinsically linked to the acquisition of high-level semantic representations.

These observations imply that autoregressive models undergo a mode switch from \textit{self-encoding} to \textit{next-prediction}. Since the prediction phase aligns with high-level semantics, we identify a clear motivation to accelerate this transition. Consequently, we adopt the REPA strategy to explicitly align intermediate model layers with pre-trained DINOv2 features. By leveraging the robust visual representations of DINOv2, we guide the model to attain the prediction-ready state earlier in the network, thereby enhancing training efficiency and generation quality.

\subsubsection{Impact of Alignment Depth on Semantic Evolution}

To investigate the influence of the alignment depth on feature evolution, \cref{fig:cknna} visualizes the CKNNA scores across layers for models aligned at depths 1, 4, 8, and 12. The vertical dashed lines indicate the specific layers where the REPA alignment is applied, while the marker size represents the magnitude of the alignment loss coefficient $\lambda$. We compute the scores at every second layer to track the progression of semantic similarity.

Qualitatively, the results demonstrate that the REPA mechanism effectively controls the semantic trajectory of the model. We observe that the layer exhibiting the highest similarity to DINOv2 features consistently shifts to coincide with the enforced alignment depth. For instance, as shown in the third subplot, when alignment is applied at Layer 8, the feature representation achieves its maximal semantic overlap with DINOv2 precisely at this layer, regardless of the loss coefficient value. This indicates that the model successfully learns to accelerate the formation of high-level semantics to match the target depth. To complement this analysis, we provide a quantitative evaluation of the performance improvements in \cref{subsec:ablation}.

\subsubsection{Ablation Studies}
\label{subsec:ablation}

\begin{wraptable}{r}{0.45\linewidth}
\vskip -0.16in
\small
  \setlength\tabcolsep{0.6mm}
  \caption{\textbf{Component analysis of LlamaGen-REPA.} All models train for 500k iterations on LlamaGen-Large. All metrics evaluate on the ImageNet validation set. ↓ indicates lower is better. Iter. denotes training iterations. Coeff. denotes the loss coefficient of the feature-alignment branch.}
  \label{tab:llamagen_repa_abla}
  \centering
  \begin{tabular}{ccccccc}
    \toprule
    \textbf{Model} & \textbf{Target Repr.} & \textbf{Depth} & \textbf{Coeff.} & \textbf{Iter.} & \textbf{gFID↓} \\
    \midrule
    LlamaGen-L & \cellcolor{red!15}- & - & - & 500k & 20.14 \\
    LlamaGen-L & \cellcolor{red!15}\textbf{DINOv2-B} & 8 & 0.5 & 500k & \textbf{17.17} \\
    \midrule
    LlamaGen-L & DINOv2-B & \cellcolor{green!15}3 & 0.5 & 500k & 18.61 \\
    LlamaGen-L & DINOv2-B & \cellcolor{green!15}4 & 0.5 & 500k & 18.57 \\
    LlamaGen-L & DINOv2-B & \cellcolor{green!15}\textbf{8} & 0.5 & 500k & \textbf{17.17} \\
    LlamaGen-L & DINOv2-B & \cellcolor{green!15}12 & 0.5 & 500k & 18.32 \\
    LlamaGen-L & DINOv2-B & \cellcolor{green!15}16 & 0.5 & 500k & 18.74 \\
    LlamaGen-L & DINOv2-B & \cellcolor{green!15}20 & 0.5 & 500k & 19.85 \\
    \midrule
    LlamaGen-L & DINOv2-B & 8 & \cellcolor{blue!15}0.5 & 500k & 17.17 \\
    LlamaGen-L & DINOv2-B & 8 & \cellcolor{blue!15}1.0 & 500k & 16.49 \\
    LlamaGen-L & DINOv2-B & 8 & \cellcolor{blue!15}\textbf{2.0} & 500k & \textbf{16.17} \\
    LlamaGen-L & DINOv2-B & 8 & \cellcolor{blue!15}3.0 & 500k & 17.30 \\
    \bottomrule
  \end{tabular}
\vskip -0.2in
\end{wraptable}
\textbf{\colorbox{red!15}{Target representation for LlamaGen-REPA.}} As shown in \cref{tab:llamagen_repa_abla}, we empirically apply the optimal REPA parameters on DiT to LlamaGen. We observe that aligning only the 8-th layer of LlamaGen-Large to the final-layer features of DINOv2-Base enables the optimal diffusion-model configuration to also accelerate convergence in the autoregressive image generation model. However, the autoregressive model must predict the next token for each token, which differs from the diffusion model that always encodes itself. Therefore, we next conduct an ablation study on which layer of the model should align with visual features.

\begin{figure*}[!t]
\centering
    \includegraphics[width=1.0\linewidth]{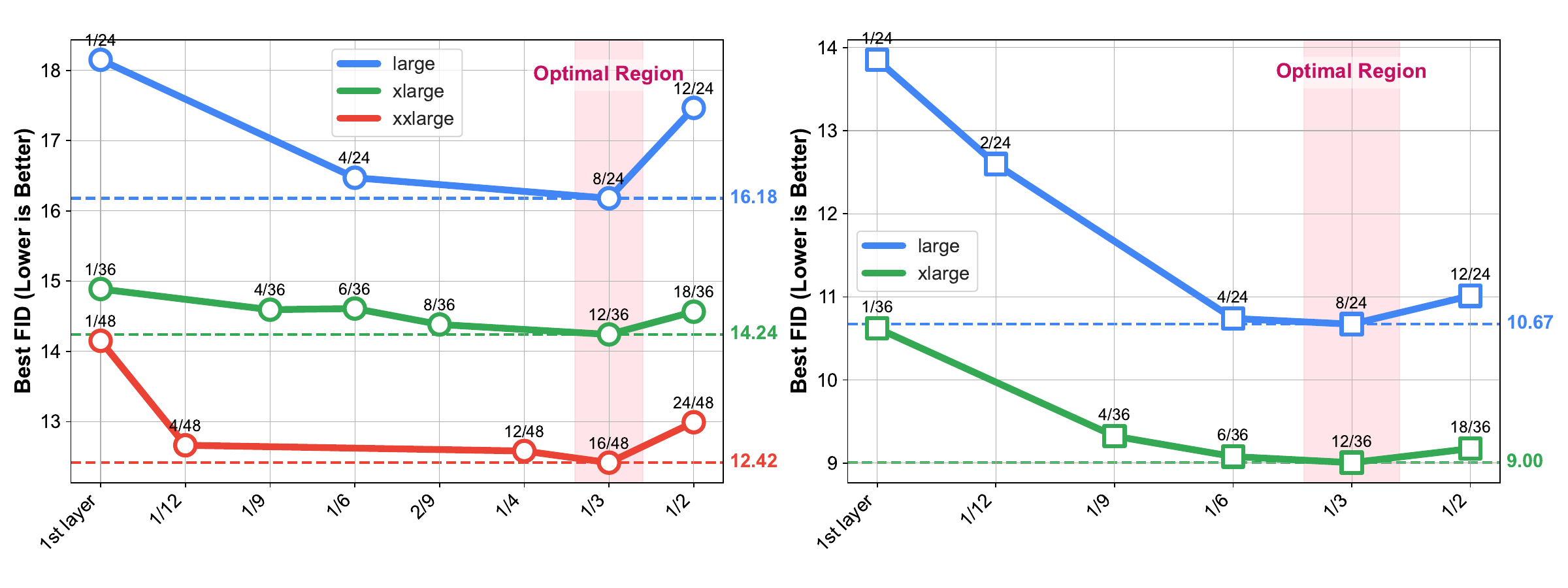} 
\caption{\textbf{Ablation study on alignment depth across different architectures and scales.} The left panel shows LlamaGen (AR) results, and the right panel shows DiT (Diffusion) results. The annotations indicate the specific alignment layer relative to the total network depth (layer/total). The optimal alignment depth consistently scales to approximately 1/3 of the total network depth (highlighted in the optimal region), rather than remaining at a fixed layer index.}
\label{fig:llamagen_dit_depth_repa}
\vskip -0.1in
\end{figure*}
\textbf{\colorbox{green!15}{Alignment depth for LlamaGen-REPA.}} In \cref{tab:llamagen_repa_abla}, we fix the coefficient of the alignment loss and perform ablations at layers 3, 4, 8, 12, 16, and 20 (out of 24 layers). We observe that performance improves with depth and peaks at layer 8. Beyond layer 8, performance gradually degrades, which we attribute to the model shifting toward next-token prediction, while per-token alignment with DINOv2 introduces a mismatch.

Additionally, to address the ambiguity regarding how REPA alignment depth generalizes across model scales, we extend the experimental validation to both autoregressive and diffusion architectures of varying sizes. \cref{fig:llamagen_dit_depth_repa} presents the best FID scores for LlamaGen (left) and DiT (right) on Large, XLarge, and XXLarge configurations.

Contrary to the assumption of a fixed optimal layer, our results reveal a proportional scaling law. While the Large model (24 layers) performs best when aligned at layer 8, this absolute value does not transfer to deeper networks. Instead, we observe that the optimal alignment depth consistently corresponds to approximately one-third of the total layers (e.g., 12/36 for XLarge and 16/48 for XXLarge). This pattern holds true for both autoregressive and diffusion paradigms, suggesting a generalized heuristic for applying REPA to larger models: alignment should be performed at the 1/3 depth mark.

\begin{wrapfigure}{r}{0.4\linewidth}
\vspace{-10pt}
\centering
\includegraphics[width=\linewidth]{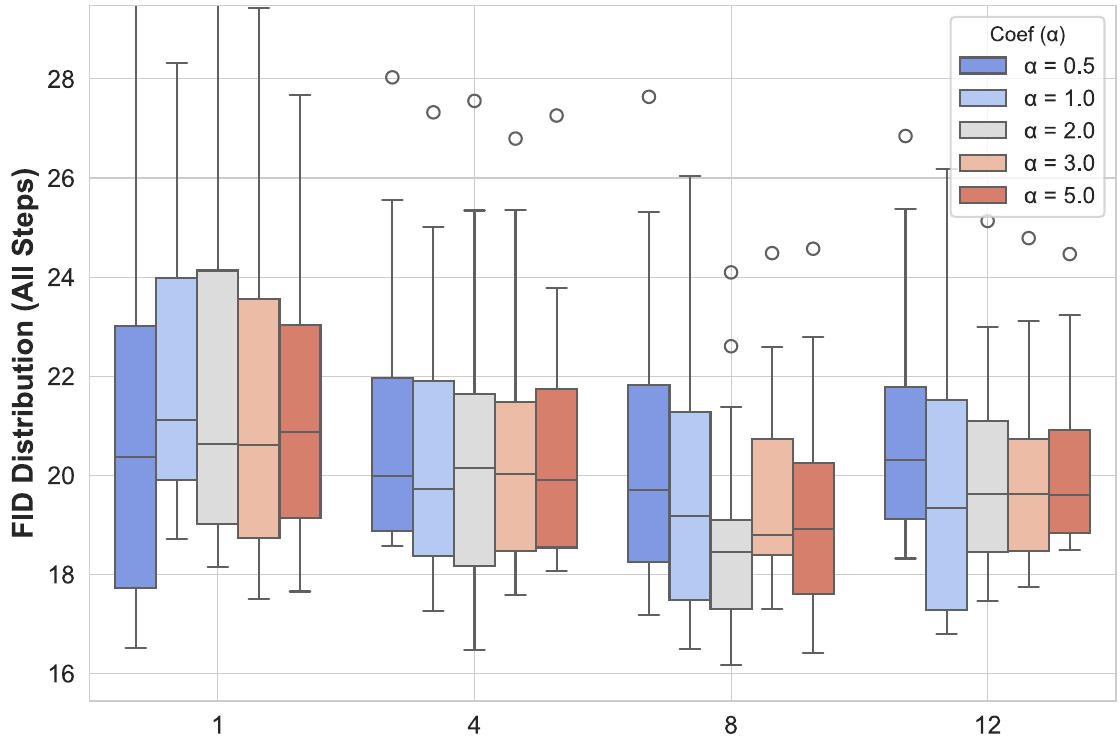}
\caption{\textbf{Extended Experiments on $\lambda$.} During training, we record all evaluation FID scores and plot them as boxplots.}
\label{fig:coef_repa_llamagen}
\vspace{-10pt}
\end{wrapfigure}
\textbf{\colorbox{blue!15}{Effect of $\lambda$ for LlamaGen-REPA.}} As shown in ~\cref{tab:llamagen_repa_abla}, we also test whether feature alignment in the AR model is sensitive to the alignment coefficient. Results show that LlamaGen-REPA benefits from a larger $\lambda$ (e.g., 2.0), which differs from DiT-REPA. We attribute this to the strong inductive bias introduced by the teacher-forcing training scheme in autoregressive models. 

Additionally, we analyze the performance of different $\lambda$ across layers in \cref{fig:coef_repa_llamagen}. Outliers typically correspond to the first checkpoint, after which the model gradually converges. Lower boxes indicate better performance, while flatter boxes indicate faster convergence. We find that $\lambda=2.0$ also achieves the best performance when the alignment depth reaches its optimum (8-th layer). Since this experiment uses a large model, the eighth layer corresponds to one-third of the total depth (24 layers), which is consistent with our earlier findings.

\section{Conclusion}

In this work, we unify the disparate paradigms of Diffusion and Autoregressive models through the lens of activation quantization. By introducing a tailored activation function, we successfully transform latent distributions into a Uniform prior, mathematically resolving the inherent trade-off between reconstruction fidelity and information efficiency. This formulation yields the \sname tokenizer, which demonstrates that the optimal equilibrium between discrete and continuous modalities resides at approximately 4 bits. Leveraging \sname as a controlled benchmark, we reveal distinct scaling behaviors: while Autoregressive models offer rapid initial convergence, Diffusion models demonstrate a superior performance ceiling. These findings suggest that the strict sequential constraints of autoregression limit ultimate generation quality.

\bibliography{iclr2025_conference}
\bibliographystyle{iclr2025_conference}
\newpage
\appendix

\newpage
\appendix
\onecolumn

\section{Background: Tokenizer for Generation}
\label{sec:bg_tok}

\textbf{Tokenizer Architecture.} The tokenizer comprises an encoder and a decoder. The encoder compresses images into a latent space, while the decoder reconstructs the latent representation back to the pixel space. Given an input image $x \in \mathbb{R}^{H \times W \times 3}$, the encoder maps it to a latent representation $z \in \mathbb{R}^{h \times w \times d}$ typically via $8\times$ or $16\times$ downsampling. Subsequently, the decoder upsamples $z$ to reconstruct the image $\hat{x}$.

\textbf{Tokenizer for Diffusion Models.} Modern diffusion models predominantly perform forward noising and reverse denoising within the latent space. Consider a latent distribution $x \sim p_{\text{data}}(x)$ and noise $\epsilon \sim \mathcal{N}(0, 1)$. By sampling a timestep $t \in [0, 1]$, the intermediate state is obtained via linear interpolation as $z_t = t x + (1-t) \epsilon$. The diffusion model predicts the velocity $v = x - \epsilon$ by minimizing the following loss function:
\begin{equation}
    \mathcal{L} = \mathbb{E}_{t, x, \epsilon} \left[ \| v_\theta (z_t, t) - v \|^2 \right]
\end{equation}
where $v_\theta$ denotes the neural network parameterized by $\theta$.

\textbf{Tokenizer for Autoregressive Models.} As autoregressive models require discrete indices, a quantization layer follows the encoder of tokenizer. This layer quantizes the latent representation $z \in \mathbb{R}^{h \times w \times d}$ into a sequence of indices $I \in \mathbb{N}^N$, where $N = h \times w$. The autoregressive model predicts the subsequent token conditioned on the preceding tokens by minimizing the cross-entropy loss:
\begin{equation}
    \mathcal{L} = - \mathbb{E}_{I} \left[ \sum_{k=1}^{N} \log p_\theta(i_k \mid i_1, \dots, i_{k-1}) \right]
\end{equation}
where $p_\theta$ represents the probability distribution predicted by the neural network parameters $\theta$, and $i_k$ denotes the $k$-th token in the index sequence.

\section{Background: Compression Ratio Analysis}
\label{sec:compression}

To theoretically quantify the efficiency of different tokenizers, we analyze their Compression Ratio (CR). We define CR as the ratio of the raw image bit-rate to the latent representation bit-rate. Consider an input image $x \in \mathbb{R}^{H \times W \times 3}$ stored in 8-bit RGB format. The total input size in bits is $S_{\text{input}} = H \cdot W \cdot 3 \cdot 8 = 24HW$.

We assume a spatial downsampling factor $f$ (typically $f=8$ or $16$), resulting in a latent spatial resolution of $h \times w$, where $h = H/f$ and $w = W/f$. The channel dimension is denoted by $d$.

\textbf{Continuous VAE:}
Standard VAEs represent latents as continuous floating-point vectors. Assuming a standard 16-bit floating-point precision (FP16 or BF16), the latent size is $S_{\text{VAE}} = h \cdot w \cdot d \cdot 16$. The compression ratio is:
\begin{equation}
    \text{CR}_{\text{VAE}} = \frac{24HW}{h \cdot w \cdot d \cdot 16} = \frac{24 f^2}{16 d} = \frac{3 f^2}{2 d}
\end{equation}
Due to the high bit-depth of floating-point numbers, VAEs typically exhibit a lower compression ratio, prioritizing reconstruction fidelity over storage efficiency.

\textbf{VQ-VAE:}
VQ-VAEs quantize the latent vector into discrete indices from a learnable codebook $\mathcal{C}$ of size $|\mathcal{C}|$ (e.g., 1024 or 8192). Each spatial location is represented by a single index, requiring $\log_2(|\mathcal{C}|)$ bits. The latent size is $S_{\text{VQ}} = h \cdot w \cdot \log_2(|\mathcal{C}|)$. The compression ratio is:
\begin{equation}
    \text{CR}_{\text{VQ}} = \frac{24HW}{h \cdot w \cdot \log_2(|\mathcal{C}|)} = \frac{24 f^2}{\log_2(|\mathcal{C}|)}
\end{equation}
VQ-VAEs achieve significantly higher compression ratios than VAEs, making them suitable for modeling long sequences in AR tasks.

\textbf{\sname (FSQ):}
FSQ does not rely on a fixed-size explicit codebook. Instead, it employs an implicit codebook defined by the number of levels $L$ per dimension $d$. The equivalent codebook size is $|\mathcal{C}| = L^d$. The total bits required to represent the scalar index $I$ is $\log_2(L^d) = d \cdot \log_2(L)$. The compression ratio is formulated as:
\begin{equation}
    \text{CR}_{\text{\sname}} = \frac{24HW}{h \cdot w \cdot d \cdot \log_2(L)} = \frac{24 f^2}{d \cdot \log_2(L)}
\end{equation}
\textbf{Comparison:} \sname (FSQ) offers a flexible trade-off. By adjusting $L$ and $d$, \sname (FSQ) can match the high compression ratio of VQ-VAE (e.g., setting $d \log_2 L \approx \log_2 |\mathcal{C}_{\text{VQ}}|$). Crucially, unlike VAEs that require 16 bits per channel, \sname (FSQ) typically requires only a few bits (e.g., $L=5 \implies \approx 2.3$ bits) per channel, yet maintains the structural properties of a continuous space before rounding.

\section{Scaling of Compression Ratio with \sname} 
\label{sec:res_cr}
\begin{figure*}[!t]
\centering
    \includegraphics[width=1.0\linewidth]{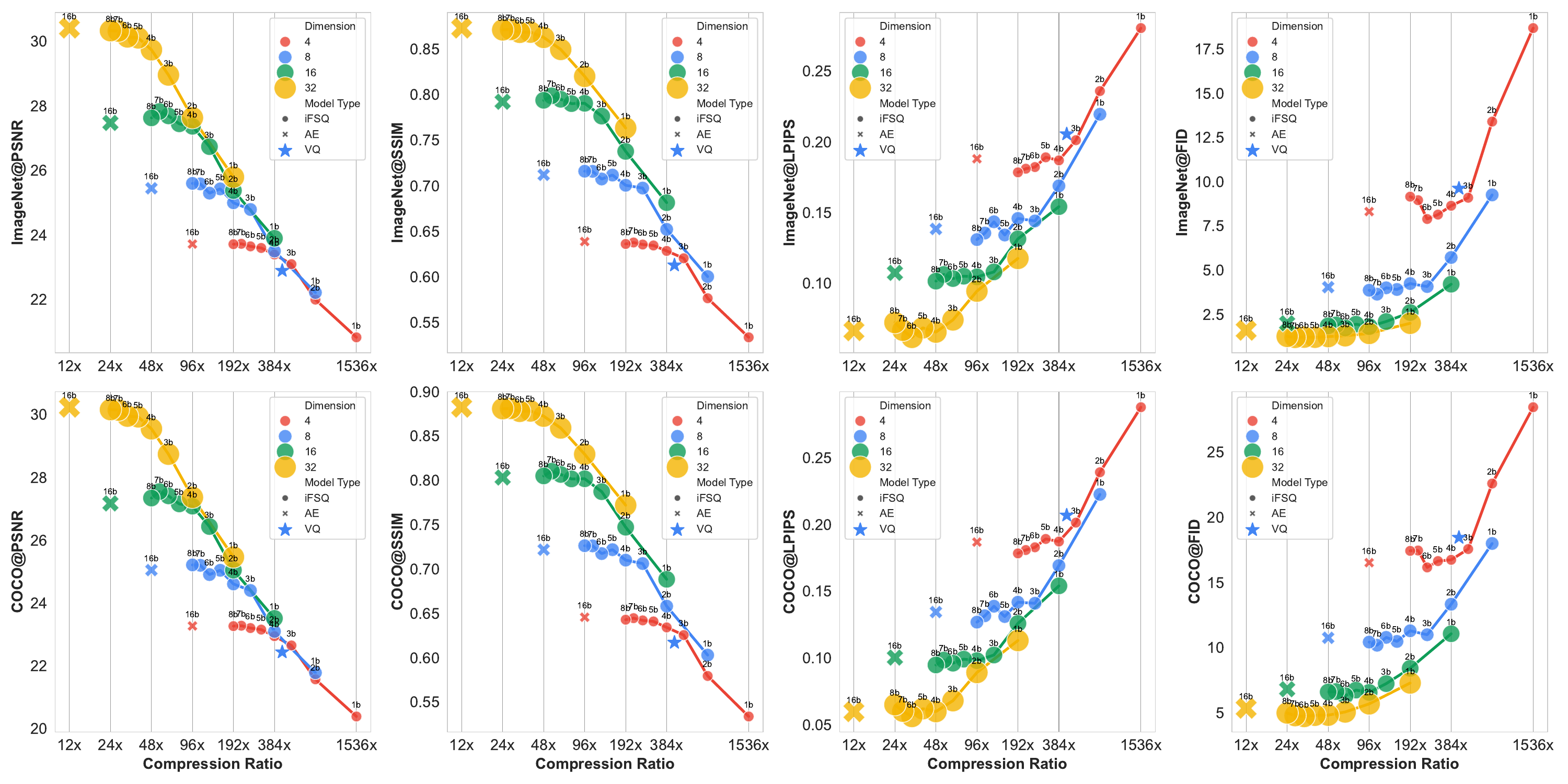} 
\caption{\textbf{Scalability of \sname.} We plot the scaling law of performance with respect to the compression ratio. Each performance point (including AE) uses a spatial compression factor of 256$\times$. The performance of AE is indicated by ``$\times$'', which train under mixed precision and use 16-bit precision to inference. The compression ratio (x-axis) is on a logarithmic scale and compression ratio of VQ ($\star$ plotted in figure) is about 438. }
\label{fig:compress_ratio}
\end{figure*}

In \cref{fig:compress_ratio}, we retrain \sname and AE under a 256$\times$ spatial compression setting to match the standard VQ configuration (16$\times$ compression in height and 16$\times$ in width). We observe that on a log-scale compression ratio, all models exhibit approximately linear performance growth or decay as compression ratio changes. A clear optimal knee point emerges around 48$\times$ compression (4 bits). We also plot VQ data points ($\star$) on the figure and find that VQ lies almost exactly on the same scaling trend, providing strong evidence that \sname serves as a compromise between discrete and continuous representations.

\section{Tokenizer Setup}
\label{sec:setup}

\begin{wraptable}{r}{0.5\linewidth}
\vspace{-4.2mm}
  \setlength\tabcolsep{1.3mm}
  \caption{\textbf{Performance comparison of tokenizer baselines.} All metrics evaluate on the ImageNet validation set. ↓ indicates lower is better. ↑ indicates higher is better.}
  \label{tab:tok_baseline}
  \centering
  \begin{tabular}{c|c|cc|cc}
    \toprule
    \textbf{Tokenizer} & \textbf{Dim} & \textbf{PSNR↑} & \textbf{SSIM↑} & \textbf{LPIPS↓} & \textbf{rFID↓} \\
    \midrule
    VQ-f8 & 8 & 26.704 & 0.770 &  \textbf{0.111} & 2.277 \\
    VQ-f8 & 4 & \textbf{26.738} & \textbf{0.780} &0.118 & \textbf{2.084} \\
    \midrule
    VQ-f16 & 8 & \textbf{22.903} & \textbf{0.613} & 0.206 & \textbf{9.626} \\
    VQ-f16 & 4 & 21.955 & 0.564 & \textbf{0.205} & 16.928 \\
    \midrule
    VAE-f8 & 4 & 27.459 & 0.799 & 0.106 & 1.998 \\
    AE-f8 & 4 & \ul{27.934} & \ul{0.813} & \ul{0.103} & \ul{1.733} \\
    AE-f8 & 8 & \textbf{30.437} & \textbf{0.878} & \textbf{0.065} & \textbf{1.661} \\
    \bottomrule
  \end{tabular}
\vspace{-3.5mm}
\end{wraptable}

\textbf{Implementation details:} For (V)AE and VQ-VAE, we follow the latent diffusion architecture. To ensure a fair comparison, all tokenizers are trained for 25 epochs on ImageNet 256$\times$256 \cite{deng2009imagenet}. LPIPS loss \cite{zhang2018unreasonable} coefficient is set to 0.1. We use the Adam optimizer \cite{kingma2014adam} with a constant learning rate of 0.001. For the model configurations of the diffusion and autoregressive models, we strictly adhere to the setup in DiT \cite{peebles2023scalable} and LlamaGen \cite{sun2024autoregressive}. To accelerate the diffusion model experiments, we use the ablation parameters from LightingDiT \cite{yao2025reconstruction}, with all diffusion models running at a batch size of 1024 and 100k iterations. The autoregressive models retain the original settings and are trained for a batch size of 256 and 500k iterations. To ensure the validity of the conclusions, all experiments are conducted on large models, such as DiT-Large or LlamaGen-Large. 

\textbf{Evaluation:} For the tokenizer reconstruction results, we report PSNR \cite{jahne2005digital}, SSIM \cite{wang2004image}, LPIPS \cite{zhang2018unreasonable}, and Fréchet Inception Distance (rFID for reconstruction) \cite{heusel2017gans}. To verify potential overfitting on ImageNet \cite{deng2009imagenet}, we additionally report reconstruction performance on COCO2017 \cite{lin2014microsoft} in \cref{fig:tanhscale}, \cref{fig:quant_level} and \cref{fig:compress_ratio}. For generated results, we report gFID (FID for generation). For diffusion models, we always use the Euler method for image generation, following LightingDiT, with the default number of function evaluations set to 250. For autoregressive models, we follow the inference settings of LlamaGen.
Since discrete tokenizers typically impose no additional distributional constraints on the latent space, whereas continuous tokenizers usually apply KL divergence regularization, we further conduct an ablation on whether to impose distributional constraints for continuous tokenizers.

\textbf{Tokenizer Baselines:} As shown in \cref{tab:tok_baseline}, we present the performance of all tokenizers under the same settings. We observe that the continuous VAE reconstruction performance is inferior to that of AE. Considering that MAETok \cite{chen2025masked} and VA-VAE \cite{yao2025reconstruction} demonstrate that the convergence of diffusion models does not depend on the constraint that the latent space maintains a standard normal distribution, we use the most standard continuous AE (without KL loss constraint) as the tokenizer for the diffusion model.

\end{document}